\documentclass[11pt]{article}
\usepackage[]{emnlp2021}
\usepackage{times,latexsym}
\usepackage{url}
\usepackage[T1]{fontenc}
\usepackage{changepage}
\usepackage{makecell}
\usepackage{xspace,mfirstuc,tabulary}
\usepackage{multirow}
\usepackage{wrapfig,lipsum,booktabs}

\newif\iftaclinstructions
\taclinstructionsfalse 
\iftaclinstructions
\renewcommand{\confidential}{}
\renewcommand{\anonsubtext}{(No author info supplied here, for consistency with
TACL-submission anonymization requirements)}
\newcommand{\instr}
\fi

\usepackage{algorithm}
\usepackage{algorithmic}
\usepackage{times}
\usepackage{latexsym}
\usepackage{amsmath}
\usepackage{amsfonts}
\usepackage{appendix}
\usepackage{mathtools}
\usepackage{amssymb}
\usepackage{bbm}
\usepackage{bm}
\usepackage{bbold}
\usepackage[inline]{enumitem}
\DeclareMathOperator*{\similarity}{sim}

\usepackage{graphicx, subfigure}
\graphicspath{ {./images/} }

\usepackage{microtype}

\title{Do Multi-Lingual Pre-trained Language Models Reveal \\Consistent Token Attributions in Different Languages?}

 \author{
 Junxiang Wang{$^\ddag$}, Xuchao Zhang{$^\dag$}, Bo Zong{$^\dag$}, Yanchi Liu{$^\dag$}, 
 \\ 
 \textbf{Wei Cheng{$^\dag$}, Jingchao Ni{$^\dag$}, Haifeng Chen{$^\dag$}, Liang Zhao{$^\ddag$}}\\ 
    
     {$\dag$}NEC Laboratories America, Princeton, NJ, USA\\
     {$\ddag$} Emory University\\
  {$^\dag$}\{xuczhang, bozong, yanchi, weicheng, jni, haifeng\}@nec-labs.com\\
  {$^\ddag$}\{junxiang.wang, liang.zhao\}@emory.edu, \\
  }

\begin{document}
\maketitle
\begin{abstract}
 During the past several years, a surge of multi-lingual Pre-trained Language Models (PLMs) has been proposed to achieve state-of-the-art performance in many cross-lingual downstream tasks. However, the understanding of why multi-lingual PLMs perform well is still an open domain. For example, it is unclear whether multi-Lingual PLMs reveal consistent token attributions in different languages. To address this, in this paper, we propose a Cross-lingual Consistency of Token Attributions (CCTA) evaluation framework.
 Extensive experiments in three downstream tasks demonstrate that multi-lingual PLMs assign significantly different attributions to multi-lingual synonyms. Moreover, we have the following observations: 1) the Spanish achieves the most consistent token attributions in different languages when it is used for training PLMs; 2) the consistency of token attributions strongly correlates with performance in downstream tasks.
\end{abstract}
\section{Introduction}
\indent The cross-lingual zero-shot transfer is a fundamental task in the NLP domain to overcome language barriers, whose goal is to transfer model information trained from source/high-resource languages (i.e. English)  to target/low-resource languages (i.e. Hindi) in the absence of explicit supervision. Multi-lingual Pre-trained Language Models (PLMs) such as multi-lingual BERT (mBERT) \cite{pires2019multilingual}, XLM \cite{conneau2019cross} and XLM-Roberta (XLM-R) \cite{conneau2020unsupervised}, have demonstrated superior performance in many cross-lingual zero-shot downstream tasks such as natural language inference and question answering. \\
\indent However, the understanding why multi-lingual PLMs perform surprisingly well is still an open domain. Previous works have investigated them extensively in various aspects. For example, they have been studied by the linguistic properties \cite{chi2020finding,edmiston2020systematic,pires2019multilingual,rama2020probing,kulmizev2020neural}, language neutrality \cite{libovicky2019language,libovicky2020language}, layer representation \cite{de2020s,singh2019bert,tenney2019bert,karthikeyan2019cross,wu2019beto}, and language generation \cite{ronnqvist2019multilingual}. Another line of related work is to understand the multi-lingual model representation in the parallel corpus \cite{kudugunta2019investigating}. They include the probing technique to investigate linguistic properties such as typological features \cite{vulic2020probing,ravishankar2019probing,ravishankar2019multilingual,bjerva2021does,bjerva2018phonology,choenni2020does,oncevay2020bridging},  and the isomorphism measure \cite{liu2019linguistic,patra2019bilingual,sogaard2018limitations,vulic2020all}.\\ \indent Even though existing literature has made much progress on the interpretation of multi-lingual PLMs, to the best of our knowledge, there still lacks an investigation on the attribution (i.e. importance) of multi-lingual tokens to the predictions of PLMs in the downstream tasks. This facilitates the understanding of how multi-lingual PLMs distinguish important tokens from others trained in source languages, and whether the understanding of tokens can be transferred to target languages. In this paper, we explore the following question in the downstream tasks, which require parallel texts (i.e. texts placed alongside their translations): \textbf{Do multi-lingual PLMs reveal consistent token attributions in different languages?} To address this, we propose a Consistency of Token Attributions (CCTA) evaluation framework. This is different from isomorphism frameworks from previous works, since they focus on the representation of tokens (i.e. token embeddings), while we focus on the importance of tokens (i.e. token attributions). Extensive experiments in three benchmark datasets (i.e. three downstream tasks) demonstrate that multi-lingual PLMs attach different attributions to multi-lingual synonyms. Moreover, we have the following observations: 1) the Spanish achieves the most consistent token attributions in different languages when it is used for training PLMs; 2) the consistency of token attributions strongly correlates with performance in downstream tasks.
\begin{figure}[t]
    \centering
    \scalebox{1}{
        \includegraphics[trim=2cm 1cm 2cm 0.2cm, width=0.92\linewidth]{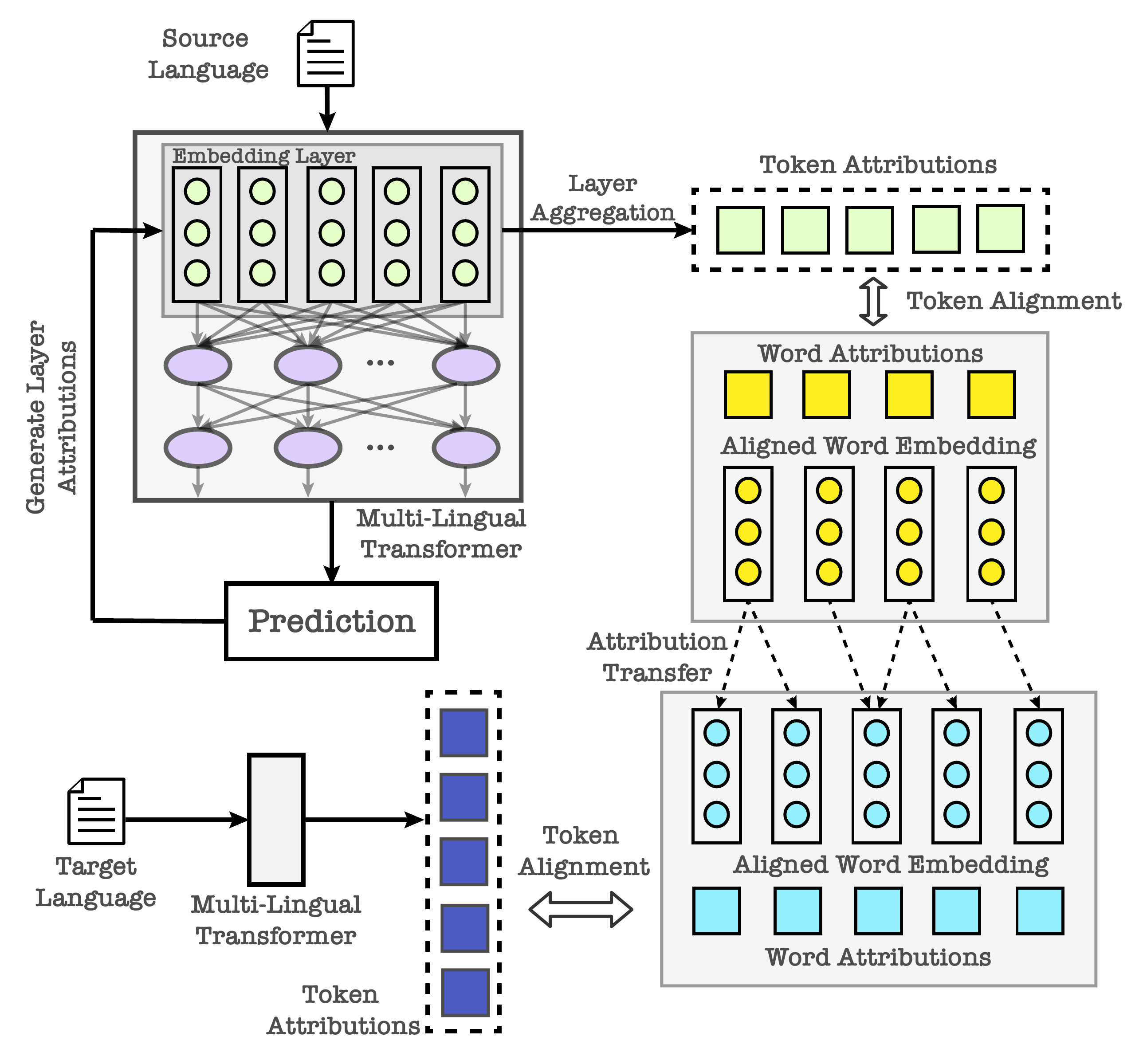}}    \caption{Overall architecture of our proposed CCTA evaluation framework.}%
    \label{fig:overall}
\end{figure} 

\section{CCTA Framework}
\label{sec:framework}
\indent This section introduces our proposed CCTA framework to evaluate the consistency of token attributions of multi-lingual PLMs in the downstream tasks, which require parallel texts. Each parallel text consists of a text and its translation from a source language and a target language, respectively. The source language is used to train multi-lingual PLMs, while the target language is used to evaluate them.  Figure \ref{fig:overall} shows the overall architecture. Firstly, we use a state-of-the-art Layer-based Integrated Gradients (LIG) method to trace token attributions. Next, we align multi-lingual token embeddings into a common comparable semantic space. Finally, the well-known earth mover's similarity is utilized to measure the consistency of token attributions via optimizing a linear programming problem. All steps are detailed in the following sections.
\subsection{Token Attribution Quantification}
\indent Given any parallel texts, the state-of-the-art Layer-based Integrated Gradients (LIG) \cite{sundararajan2017axiomatic} is applied to quantify token attributions. In contrast with previous attribution methods, LIG follows the axioms of attribution methods and tease apart errors from the misbehaviors of multi-lingual PLMs. It measures how the input gradient is changed by a relative path, and therefore needs a reference (i.e. baseline). Given an input vector $x\in \mathbb{R}^d$, its baseline $x{'} \in \mathbb{R}^d$ (i.e. the starting point of path $x{'}\rightarrow x$), and a prediction function $F$,  the change of gradient $\frac{\partial F}{\partial x}$  along the path $x{'}\rightarrow x$ is shown as follows:
\begin{align}
    \rm{LIG}_i(x)\!=\!(x\!-\!x')\times\int_{\alpha=0}^{1}\frac{\partial F(x'\!+\!\alpha(x\!-\!x'))}{\partial x}d\alpha \label{eq:IG}
\end{align}
where $x_i$ is the $i$-th dimension of $x$. Obviously, as $\alpha$ increases from $0$ to $1$, the path starts from $x'$ to $x$, and LIG integrates the gradient $\frac{\partial F}{\partial x}$ along the path. Equation \eqref{eq:IG} requires the differentiability of  $F$. Unfortunately, the input of  a multi-lingual PLM is a sequence of non-differentiable token IDs. To address this issue, the embedding layer of a multi-lingual PLM is chosen to be an origin as the input and all embedding attributions are aggregated. 
The baseline in Equation \eqref{eq:IG} is chosen as follows: we leave separation tokens and replace other tokens with  padding tokens in any sentence. Let $d$ be the dimensionality of the embedding layer, given a parallel text $(s,t)$,  where  $s_i$ and $t_j$ are the $i$-th  and  $j$-th tokens of sentences $s$ and $t$, respectively, attributions are aggregated mathematically as follows:
\begin{align*}
    w_{s_i}=\sum\nolimits_{k=1}^d \rm{LIG}_k(s_i), \ w_{t_j}=\sum\nolimits_{k=1}^d \rm{LIG}_k(t_j)
\end{align*}
where $w_{s_i}$, and $w_{t_j}$ are attributions of $s_i$ and $t_j$, respectively. Namely, the attributions of tokens $s_i$ and $t_j$ are the sum of their attributions along the dimensionality of the embedding layer.
\subsection{Multi-Lingual Aligned Token Embeddings}
\label{sec:Multi-Lingual Aligned Token Embeddings}
\indent Multi-lingual PLMs usually provide contextual embeddings, which are mapped in different semantic spaces \cite{peters2018deep}. In order to bridge the semantic gap, token embeddings are aligned to a shared context-free semantic space. Suppose $e_{s_i}$ and $e_{t_j}$ are denoted as embeddings of $s_i$ and $t_j$ in such a shared semantic space, respectively, then the semantic similarity between them is measured by the cosine similarity, which is shown as follows:
\begin{align*}
    \similarity(s_i,t_j)=\frac{{e_{s_i}}^T e_{t_j}}{\Vert e_{s_i}\Vert\Vert e_{t_j}\Vert}
\end{align*}
\subsection{Consistency of Token Attributions}
\label{sec:Consistency of Interpretations Measure}
\indent Finally, the well-known Earth mover's similarity is used to measure the consistency of token attributions between a source language and a target language \cite{hitchcock1941distribution}. It is obtained by optimizing a linear programming problem as follows:
\begin{align}
    &C(s,t)\!=\!\max \sum_{i=1,s_i\in s}^l\! \sum_{j=1,t_j\in t}^l f_{i,j}\similarity(s_i,t_j) \label{eq:problem}\\
    & \nonumber s.t. \ \sum\nolimits_{j=1}^l f_{i,j}\leq \overline{w}_{s_i} \ (i=1,\cdots,l)\\
    & \nonumber \sum\nolimits_{i=1}^l f_{i,j}\leq \overline{w}_{t_j} \ (j=1,\cdots,l), \ f_{i,j}\geq 0
\end{align}
where $C(s,t)$ is the consistency of token attributions, and $l$ is the maximal length of sentences $s$ and $t$. $\overline{w}_{s_i}$ and $\overline{w}_{t_j}$ are denoted as the normalized values of $w_{s_i}$ and $w_{t_j}$, respectively, or namely $\sum\nolimits_{s_i\in s}\overline{w}_{s_i} = 1,\overline{w}_{s_i}\geq 0, \sum\nolimits_{t_j\in t}\overline{w}_{t_j} = 1, \overline{w}_{t_j}\geq 0$. The weight $f_{i,j}$ quantifies the consistency of token attributions from $s_i$ to $t_j$. The larger $C(s,t)$ is, the more likely multi-lingual PLMs attach equal importance to multi-lingual synonyms. Equation \eqref{eq:problem} can be efficiently optimized by an existing linear programming solver \cite{mitchell2011pulp}.

\section{Experiments}
\label{sec:experiment}
\indent In this section, we evaluate our proposed CCTA framework in three datasets. All experiments were conducted on a Linux server with the
Nvidia Quadro RTX 6000 GPU and 24GB memory \footnote{The code and the empirical study are available in the supplementary materials.}.
 \subsection{Datasets and Models}
 \indent The CCTA evaluation framework was tested in three datasets (i.e. three downstream tasks): XNLI, PAWS-X, and XQuAD. XNLI is a benchmark dataset used for the cross-lingual natural language inference task, and 13 languages were studied in the XNLI: English (en), Arabic (ar), Bulgarian (bg), German (de), Greek (el), Spanish (es), French (fr), Hindi (hi), Russian (ru),  Thai (th), Turkish (tr), Vietnamese (vi), and Chinese (zh) \cite{hu2020xtreme}. PAWS-X is a benchmark dataset used for the cross-lingual paraphrase identification task, and six languages were studied in the PAWS-X: English (en), German (de), Spanish (es), France (fr), Korea (ko), and Chinese (zh) \cite{hu2020xtreme}. XQuAD is a benchmark dataset used for the cross-lingual question answering task, and 11 languages were explored in the XQuAD: English (en), Arabic (ar), German (de), Greek (el), Spanish (es), Hindi (hi),  Russian (ru), Thai (th), Turkish (tr), Vietnamese (vi) and Chinese (zh) \cite{hu2020xtreme}.\\
\indent Three state-of-the-art multi-lingual PLMs are utilized for comparison: multi-lingual BERT (mBERT), Cross-lingual Language Model (XLM), and XLM-Roberta (XLM-R). The XLM and the XLM-R both have a base model and a large model. We fine-tuned all models for 10 epochs trained in English and selected the best models based on the performance of the built-in dev sets of all languages in three datasets. Finally, the performance was evaluated in the test set of all languages in three datasets. The Adam optimizer was used with a learning rate of 1e-5 and no weight decay \cite{kingma2015adam}. The batch sizes for the base model and the large model of the XLM and the XLM-R were set to 128 and 32, respectively, and the batch size for the mBERT was set to 32. All models were pretrained using Masked Language Modeling (MLM).
  \begin{table*}[]
\scriptsize
     \centering
     \caption{All consistency scores of token attributions in the XNLI dataset.}
     \label{tab:XNLI consistency}
     \begin{tabular}{c|c|c|c|c|c|c|c|c|c|c|c|c|c|c}
         \hline\hline Model&en&ar&bg&de&el&es&fr&hi&ru&th&tr&vi&zh&Overall  \\\hline
         mBERT&1.000&0.194&0.240&0.270&0.245&0.331&0.290&0.217&0.229&0.207&0.245&0.232&0.336&0.310\\\hline
          XLM-Base&1.000&0.206&0.256&0.291&0.257&0.352&0.310&0.219&0.249&0.214&0.262&0.243&0.352&0.324
          \\\hline
          XLM-Large&1.000&0.198&0.245&0.273&0.250&0.334&0.294&0.222&0.232&0.211&0.248& 0.235&0.340&0.314\\\hline
          XLM-R-Base&1.000&0.195&0.242&0.270&0.246&0.330&0.290&0.219&0.230&0.210&0.244&0.233&0.335&0.311\\\hline
          XLM-R-Large&1.000&0.198&0.245&0.275&0.248&0.334&0.294&0.222&0.234&0.212&0.248&0.234&0.342&0.314\\\hline\hline
     \end{tabular}
 
     \centering
     \caption{All consistency scores of token attributions in the PAWS-X dataset. }
     \label{tab:pawsx consistency}
     \begin{tabular}{c|c|c|c|c|c|c|c}
         \hline\hline Model&en&de&es&fr&ko&zh&Overall  \\\hline
          mBERT&1.000&0.411&0.459&0.423&0.266&0.337&0.483
          \\\hline
          XLM-Base&1.000&0.416&0.465&0.429&0.272&0.339&0.487\\\hline
          XLM-R-Base&1.000&0.406&0.455&0.418&0.262&0.333&0.479\\\hline\hline
     \end{tabular}
     
   \scriptsize
     \centering
     \caption{All consistency scores of token attributions in the XQuAD dataset.}
     \label{tab:xquad consistency}
     \begin{tabular}{c|c|c|c|c|c|c|c|c|c|c|c|c}
         \hline\hline
         \multicolumn{11}{c}{Start Position of Answer}\\\hline
         Model&en&ar&de&el&es&hi&ru&th&tr&vi&zh&Overall  \\\hline
          mBERT&1.000&0.234&0.349&0.299&0.414&0.276&0.285&0.237&0.321&0.279&0.363&0.369
          \\\hline
          XLM-R-Base&1.000&0.233&0.343&0.295&0.409&0.275&0.282&0.235&0.319&0.279&0.356&0.366\\\hline
            \multicolumn{11}{c}{End Position of Answer}\\\hline
Model&en&ar&de&el&es&hi&ru&th&tr&vi&zh&Overall  \\\hline
          mBERT&1.000&0.235&0.350&0.299&0.415&0.278&0.285&0.237&0.320&0.281&0.364&0.369
          \\\hline
          XLM-R-Base&1.000&0.235&0.345&0.297&0.411&0.277&0.283&0.237&0.319&0.281&0.357&0.367\\\hline\hline
     \end{tabular}
 \end{table*}
 \subsection{Inconsistent Token Attributions}
 Tables \ref{tab:XNLI consistency}, \ref{tab:pawsx consistency} and \ref{tab:xquad consistency} demonstrate the consistency of token attributions between English and all languages in the XNLI, the PAWS-X, and the XQuAD, respectively. Obviously,  most of the scores are below 0.5. For example, the best scores aside from English are around 0.35, 0.46, and 0.36 in the XNLI, the PAWS-X, and the XQuAD, respectively. This indicates that multi-lingual PLMs assign different attributions to multi-lingual synonyms. Moreover, while little distinctions are shown among multi-lingual PLMs, some gaps are found among languages. For example, the scores of all multi-lingual PLMs in zh (Chinese) are about 0.14 higher than these in ar (Arabic) in the XNLI.

\subsection{Most Consistent Token Attributions Trained in Spanish}
\begin{table}[]
     \centering
     \normalsize
     \caption{The test accuracy in the PAWS-X dataset for different source languages using the XLM-R-Base model. }
     \label{tab:pawsx-not-english performance}
     \tiny
     \begin{tabular}{p{0.9cm}|p{0.5cm}|p{0.5cm}|p{0.5cm}|p{0.5cm}|p{0.5cm}|p{0.5cm}|p{0.7cm}}
         \hline\hline Language&en&de&es&fr&ko&zh&Overall  \\\hline
        en&0.944&0.868&0.887&0.882&0.732&0.799&0.852\\\hline
          de&0.936&0.878&0.888&0.893&0.758&0.808&0.860\\\hline
          es&0.935&0.875&0.901&0.899&0.774&0.808&\textbf{0.865}\\\hline
          fr&0.933&0.875&0.897&0.901&0.753&0.814&0.862\\\hline ko&0.907&0.854&0.859&0.862&0.807&0.828&0.853\\\hline zh&0.922&0.856&0.870&0.861&0.790
&0.837&0.860\\\hline
\hline
     \end{tabular}
     \centering
     \normalsize
     \caption{All consistency scores of token attributions in the PAWS-X dataset for different source languages using the XLM-R-Base model. }
     \label{tab:pawsx-not-english consistency}
     \tiny
     \begin{tabular}{p{0.9cm}|p{0.5cm}|p{0.5cm}|p{0.5cm}|p{0.5cm}|p{0.5cm}|p{0.5cm}|p{0.7cm}}
         \hline\hline Language&en&de&es&fr&ko&zh&Overall  \\\hline
        en&1&0.406&0.455&0.418&0.262&0.333&0.479\\\hline
          de&0.410&1&0.513&0.504&0.392&0.294&0.519\\\hline
          es&0.458&0.512&1&0.604&0.424&0.341&\textbf{0.557}\\\hline
          fr&0.419&0.502&0.602&1&0.410&0.315&0.541\\\hline ko&0.263&0.390&0.425&0.411&1&0.302&0.465\\\hline zh&0.331&0.291&0.340&0.315&0.302&1&0.430\\\hline\hline
     \end{tabular}
 \end{table}

\indent Next, Tables \ref{tab:pawsx-not-english performance} and \ref{tab:pawsx-not-english consistency} demonstrate the test accuracy and consistency scores of token attributions in the PAWS-X dataset for different source languages, respectively. Every row and column represent a source language and a target language, respectively. Spanish (es) achieves the most consistent token attributions: it not only performs well in close languages such as English (en), German (de), and French (fr), it also reaches a fair score in distant languages such as Korea (ko). The scores in French (fr) and German (de) are also better than the score in English.
\subsection{Strong Correlations Between Performance and Consistency Scores}
     \begin{figure}
             \includegraphics[width=\columnwidth]{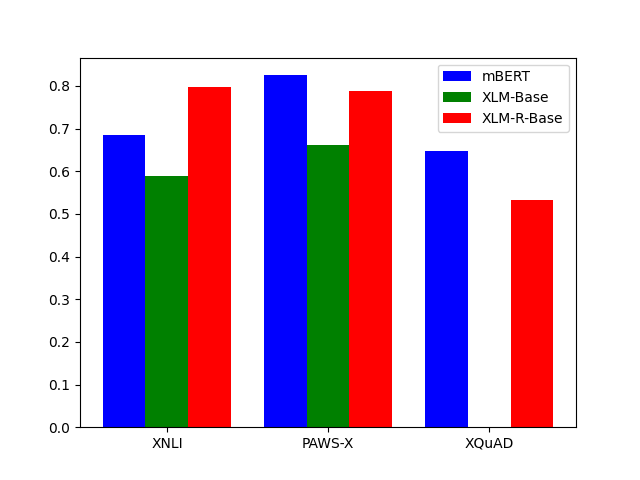}
                  \caption{The correlation coefficients between performance and consistency scores in three datasets (The correlation coefficient of the XLM-Base model in the XQuAD dataset is unavailable).}
                  \label{fig:correlation}
     \end{figure}

\indent Finally, the correlation coefficients between performance and consistency scores in three datasets and different multi-lingual PLMs are shown in Figure \ref{fig:correlation}. It indicates that all multi-lingual PLMs demonstrate strong positive correlations. Moreover, the correlation coincides with the difficulty of the dataset: the simpler a dataset is, the stronger correlation it has. 
\section{Conclusion}
\label{sec:conclusion}
\indent In this work, we propose a CCTA framework to assess the consistency of token attributions of multi-lingual PLMs. Specifically, given parallel texts, token attributions (i.e. importance) are quantified by the state-of-the-art Layer-based Integrated Gradients (LIG) method. Then all tokens are aligned into a common comparable embedding space. Finally, the well-known earth mover's similarity is utilized to measure the consistency scores. Experimental results in three downstream tasks show that  PLMs achieve inconsistent token attributions.

\clearpage
\bibliographystyle{acl_natbib}
\bibliography{tacl2018}
\end{document}


\appendix
\onecolumn
\textbf{Appendix}\\
 \begin{table*}[!hbp]
         \centering
         \caption{Two parallel texts in English and Chinese in the XNLI dataset.}
         \label{tab:xnli case study}
         \begin{tabular}{c|p{7cm}|p{7cm}}
         \hline\hline
          \multicolumn{3}{c}{\begin{makecell}
                   {The parallel text with the highest consistency score of token attributions (contradiction)}
          \end{makecell}}\\\hline
              Language&Premise&Hypothesis  \\\hline
              English& H. H. Richardson and his protege Charles Follen McKim were alumni, as well as McKim's assistants, John M. Carrare and Thomas Hastings.  & Charles Follen McKim, John M. Carrare and Thomas Hastings were all alumni, but H. H. Richardson was not and was from another university.\\\hline
              Chinese&
              H. H. Richardson和他的门徒Charles Follen McKim是校友，还有McKim的助手John M. Carrare和Thomas Hastings。&
              Charles Follen McKim、John M. Carrare和Thomas Hastings都是校友，但H. H. Richardson则不是，他来自另一所大学。\\
              \hline\hline
          \multicolumn{3}{c}{\begin{makecell}
                   {The parallel text with the lowest consistency score of token attributions (contradiction)}
          \end{makecell}}\\\hline
          Language&Premise&Hypothesis  \\\hline
          English&You know who would understand?& Do you know who won't understand?    \\\hline
          Chinese&你知道谁会明白吗？&
你知道还有谁不明白吗？\\\hline\hline
         \end{tabular}
         \centering
         \normalsize
         \caption{ Attributions of two English and Chinese texts in the XNLI dataset.}
         \label{tab:xnli attribution}
         \begin{tabular}{c|c|c|c}
         \hline\hline
         \multicolumn{4}{c}{\begin{makecell}
                   {The parallel text with the highest consistency score of token attributions (contradiction)}
          \end{makecell}}\\\hline
         \multicolumn{4}{c}{Premise}\\\hline
         \begin{makecell} {English}\end{makecell} &\begin{makecell} {Attribution}\end{makecell} &\begin{makecell} {Chinese}\end{makecell}&\begin{makecell} {Attribution}\end{makecell}\\\hline
              H. H. Richardson&0.1&H. H. Richardson&0.11\\\hline
              and&0.02&和&0.02\\\hline
              his protege&0.04&他的门徒&0.04\\\hline
              Charles Follen McKim&0.04 &Charles Follen McKim &0.05\\\hline
              were&0.02&是&0.01\\\hline
              alumni&0.02&校友&0.02\\\hline
              as well as&0.05&还有&0.01\\\hline
              McKim's assistants&0.04&McKim的助手&0.06\\\hline
              John M. Carrare&0.07&John M. Carrare&0.07\\\hline
              and&0.02&和&0.02\\\hline
              Thomas Hastings&0.03&Thomas Hastings&0.04\\\hline
              \multicolumn{4}{c}{Hypothesis}\\\hline
              \begin{makecell} {English}\end{makecell} &\begin{makecell} {Attribution}\end{makecell} &\begin{makecell} {Chinese}\end{makecell}&\begin{makecell} {Attribution}\end{makecell}\\\hline
              Charles Follen McKim&0.04&Charles Follen McKim&0.05\\\hline
              John M. Carrare&0.06&John M. Carrare&0.09\\\hline
              and&0.02&和&0.02\\\hline
              Thomas Hastings&0.02&Thomas Hastings&0.03\\\hline
              were all&0.02&都是&0.01\\\hline
              alumni&0.02&校友&0.01\\\hline
              but&0.02&但&0.03\\\hline
              H. H. Richardson was not&0.14&H. H. Richardson则不是&0.17\\\hline
              and was from another university&0.09&他来自另一所大学&0.1\\
         \hline\hline
                 \multicolumn{4}{c}{\begin{makecell}
                   {The parallel text with the lowest consistency score of token attributions (contradiction)}
          \end{makecell}}\\\hline
         \multicolumn{4}{c}{Premise}\\\hline
         \begin{makecell} {English}\end{makecell} &\begin{makecell} {Attribution}\end{makecell} &\begin{makecell} {Chinese}\end{makecell}&\begin{makecell} {Attribution}\end{makecell}\\\hline
         You know&0.04&你知道&0.14\\\hline
         who would understand&0.04&谁会明白吗&0.15\\\hline
              \multicolumn{4}{c}{Hypothesis}\\\hline
              \begin{makecell} {English}\end{makecell} &\begin{makecell} {Attribution}\end{makecell} &\begin{makecell} {Chinese}\end{makecell}&\begin{makecell} {Attribution}\end{makecell}\\\hline
              Do you know&0.12&你知道&0.15\\\hline
              who won't understand&0.75&还有谁不明白吗&0.43\\\hline\hline
         \end{tabular}
     \end{table*}
     \indent In this section, we conduct some empirical studies in order to illustrate the effectiveness of the CCTA evaluation framework. Two parallel texts of English and Chinese in the XNLI dataset are shown in Table \ref{tab:xnli case study}. They represent the most and the least consistency score of token attributions, respectively. All phrases and corresponding normalized attributions are shown in Table \ref{tab:xnli attribution}. Some punctuation marks also have normalized attributions, and are not shown in Table \ref{tab:xnli attribution}.  As shown in Table \ref{tab:xnli case study}, the contradiction between the premise and the hypothesis of the first parallel text consists in whether H.H.Richardson and other persons were alumni, and in Table \ref{tab:xnli attribution} English phrases such as "
H. H. Richardson", "H. H. Richardson was not" and "and was from another university" and their Chinese counterparts receive higher attributions than other tokens (all are above 0.09). This indicates not only the consistency scores of the English sample are reasonable, but also show a excellent consistency between English and Chinese. Moreover, normalized attributions between every English token and its Chinese counterpart are very close. As for the second parallel text, "who would understand" in the premise contradicts "who won't understand" in the hypothesis. However, there is a clear distinction between the consistency scores of the English example and its Chinese counterpart: while the English token "Who won't understand" and its Chinese counterpart does play the most important role in the interpretation, but more attributions in the Chinese sample are attached in the premise, which leads to inconsistent token attributions.